\documentclass[letterpaper,journal]{IEEEtran}
\usepackage{amsmath,amsfonts,amssymb}
\usepackage{algorithmic}
\usepackage{algorithm}
\usepackage{array}
\usepackage[caption=false,font=normalsize,labelfont=sf,textfont=sf]{subfig}
\usepackage{textcomp}
\usepackage{stfloats}
\usepackage{url}
\usepackage{verbatim}
\usepackage{graphicx}
\usepackage{cite}
\usepackage{bm}
\usepackage{booktabs}
\usepackage{multirow}
\usepackage{siunitx}

\usepackage{xcolor}

\begin{document}

\title{AUSLUN: A Fixed-Hover UAV--USV System for GNSS-Denied Maritime Search and Navigation}

\author{Siyuan Yang, Zikai Jia, Hailiang Kuang, Xiaoyu He, Qizhi Guo, Yihao Dong\IEEEauthorrefmark{1} and Shaoming He\IEEEauthorrefmark{1}
	\thanks{S. Yang is with the School of Aerospace Engineering, Beijing Institute of Technology, Beijing 100081, China (e-mail: siyuan.yang@bit.edu.cn).}
	\thanks{Z. Jia is with the School of Aerospace Engineering, Beijing Institute of Technology, Beijing 100081, China (e-mail: 3120215032@bit.edu.cn).}
	\thanks{H. Kuang is with the School of Aerospace Engineering, Beijing Institute of Technology, Beijing 100081, China (e-mail: cpt.kk@live.com).}
	\thanks{X. He is with the School of Aerospace Engineering, Beijing Institute of Technology, Beijing 100081, China (e-mail: 3120230048@bit.edu.cn).}
	\thanks{Q. Guo is with the School of Aerospace Engineering, Beijing Institute of Technology, Beijing 100081, China (e-mail: qizhi.guo@bit.edu.cn).}
	\thanks{Y. Dong is with the Khalifa University Center for Autonomous Robotic Systems (KUCARS), Khalifa University, Abu Dhabi 127788, United Arab Emirates (e-mail: yihao.dong@ku.ac.ae).}
	\thanks{S. He is with the School of Aerospace Engineering, Beijing Institute of Technology, Beijing 100081, China (e-mail: shaoming.he@bit.edu.cn).}
	\thanks{\IEEEauthorrefmark{1}Corresponding authors.}}

\markboth{}{Yang \MakeLowercase{\textit{et al.}}: AUSLUN}

\maketitle

\begin{abstract}
Global navigation satellite system (GNSS) denial can prevent an unmanned surface vehicle (USV)
from both finding a distant vessel and maintaining a globally referenced approach.
This paper presents AUSLUN (Automatic UAV Search, Localization, and USV Navigation),
a fixed-hover aerial-surface system that uses a coastal unmanned aerial vehicle (UAV),
which estimates its own pose through visual-inertial odometry (VIO),
as a long-range sensing and navigation anchor.
The central design shifts sensing motion from UAV translation to a zoom pod
and closes the loop through three coupled elements:
polygon-aware annular pod scanning,
modality-aware bearing-range localization,
and target-relative USV guidance with visual-loss recovery.
The same gated recursive estimator uses laser range for the non-cooperative target
and datalink range for the cooperative USV.
Search-planning simulations show that the adaptive yaw bounds reduce scan time and redundant coverage relative to a matched fixed-sector scan,
and GPS-referenced field data show that the gated recursive estimator outperforms non-recursive baselines in localization accuracy.
An integrated maritime mission further demonstrates the complete search-to-navigation sequence,
including a deliberately triggered visual-loss recovery.
These results establish the feasibility and operating boundary of fixed-hover UAV assistance
for stationary-target approach in coastal GNSS-denied environments.
The source code and a video demonstration are publicly available\footnote{Code: \url{https://github.com/xirhxq/pod_search}; Video demonstration: \url{https://youtu.be/S-5RkJs35JI}}.
\end{abstract}


\section{Introduction}
\label{sec:introduction}

\begin{figure}[t]
\centering
\includegraphics[width=\columnwidth]{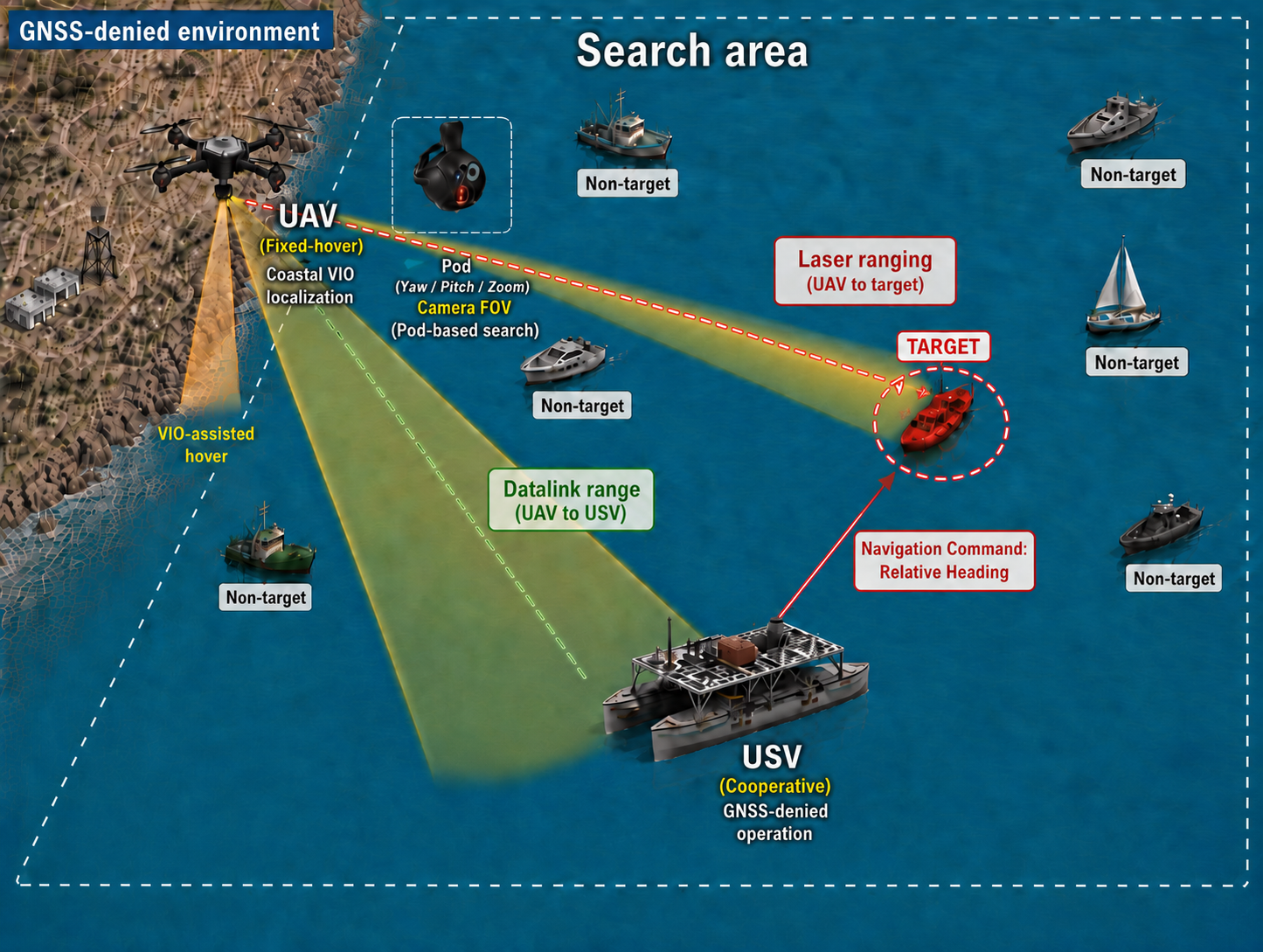}
\caption{Fixed-hover UAV assistance for GNSS-denied maritime target approach.
The UAV remains over a coastal VIO-feasible location,
searches the water region by pod motion,
uses laser and datalink ranges for the non-cooperative target and cooperative USV,
respectively,
and transmits target-relative guidance to the USV.}
\label{fig:overview}
\end{figure}

Unmanned surface vehicles (USVs) are increasingly used for maritime surveillance,
asset protection,
search and rescue,
and harbor operations \cite{mahacek2012dynamic,villa2020path}.
A demanding part of these missions is long-range target approach:
the USV must identify and approach one vessel among surrounding traffic
before onboard vision or light detection and ranging (LiDAR) can provide reliable close-range acquisition.
The task becomes more difficult when global navigation satellite system (GNSS) measurements
are degraded by blockage,
multipath,
interference,
or spoofing \cite{meng2025trusted}.
The resulting problem is simultaneously perceptual and navigational:
the target is outside the USV sensing horizon,
while the USV lacks a stable global reference for the approach.
This paper studies the fixed-hover aerial-assistance configuration in Fig.~\ref{fig:overview}.

The key systems insight is to retain the UAV over a low-altitude coastal location
where shoreline features support VIO
\cite{shan_lvi-sam_2021,weinstein_vio_2018},
and to transfer the required sensing motion to a pitch-yaw-zoom pod.
The fixed hover point then acts as a persistent reference
for long-range bearing-range measurements.
This arrangement differs from conventional coverage planning,
which primarily moves the vehicle through the search region
\cite{paull2013sensor,chen2020active},
and from visual servoing,
which usually treats visibility as a constraint on a single tracking task
\cite{xie2017input,zheng2019visibility,li2023robocentric,qin2023perception}.
Here,
pod search,
non-cooperative target localization,
cooperative USV localization,
guidance,
and recovery must operate as one closed mechatronic loop.

Three couplings make this loop nontrivial.
First,
a fixed viewpoint must cover a polygonal water region using only pod pitch,
yaw,
and field of view (FOV),
while maintaining sufficient apparent target size and view overlap.
Second,
the two vessels share the same bearing geometry but not the same range source:
the target is non-cooperative and requires visual alignment before laser ranging,
whereas the USV supplies a quantized datalink range.
Third,
asynchronous sensing and intermittent visual tracking must be converted into
continuous target-relative guidance without accepting invalid bearing-range pairs.
These couplings are not resolved by a detector,
coverage planner,
relative-localization estimator,
or guidance law in isolation.

We therefore develop AUSLUN (Automatic UAV Search, Localization, and USV Navigation),
a supervisory fixed-hover UAV--USV system.
A polygon-to-annular planner converts the mission region into FOV-aware pod sweeps with adaptive yaw bounds.
A gated recursive estimator switches the range modality according to vessel cooperation
while retaining a common bearing-range state model.
The resulting target and USV states are converted into relative guidance,
with separate transitions for prolonged target loss,
local USV reacquisition,
and close-range LiDAR handoff.
The framework was deployed during the Mohamed Bin Zayed International Robotics Challenge (MBZIRC) 2023
\cite{mbzirc2023}
and evaluated using search-planning simulations,
an integrated Yas Island mission,
a deliberately activated recovery interval,
and a Global Positioning System (GPS)-referenced localization segment.

The main contributions of this work are as follows.
\begin{enumerate}
    \item \emph{Fixed-hover search formulation and pod planner:}
    We formulate maritime coverage under a VIO-constrained fixed UAV position
    and generate multi-ring pod scans whose pitch,
    FOV,
    and polygon-dependent yaw bounds jointly enforce detectability and overlap requirements.
    Relative to a matched fixed-sector scan,
    the planner reduces planned search time by 10.9--55.6\% across three region geometries.

    \item \emph{Modality-aware bearing-range localization:}
    We develop a gated estimation architecture that uses a common vessel state model
    with laser range for the non-cooperative target
    and datalink range for the cooperative USV.
    The gating logic explicitly couples visual confidence,
    pod settling,
    and asynchronous range validity;
    the field segment yields \SI{11.4}{\meter} mean error
    and \SI{16.0}{\meter} 95th-percentile error.

    \item \emph{Field-verified aerial-surface supervisory loop:}
    We integrate search,
    localization,
    target-relative guidance,
    local visual-loss recovery,
    and USV LiDAR handoff in one UAV-side state machine.
    Maritime field data demonstrate the complete sequence
    and delineate its operating boundary for stationary targets,
    coastal line of sight,
    and a VIO-feasible hover point.
\end{enumerate}

The remainder of the paper positions AUSLUN relative to coverage,
cooperative localization,
and maritime perception;
defines the constrained aerial-surface problem;
describes the planning,
estimation,
and supervisory algorithms;
and evaluates the resulting system in simulation and field experiments.

\section{Related Work}
\label{sec:related}

The proposed system lies at the intersection of sensor-aware coverage,
bearing-range localization,
and aerial-surface cooperation.
The relevant distinction is therefore not whether each component has appeared before,
but whether prior work closes the same perception-estimation-guidance loop
under a fixed coastal viewpoint.

\subsection{Sensor-Aware Coverage and Visual Servoing}

Coverage planning has been studied for ground,
marine,
and aerial robots \cite{galceran2013survey},
including sensor-driven underwater inspection,
active simultaneous localization and mapping,
camera-footprint-aware aerial coverage,
zoom-aware planning,
and integrated gimbal guidance
\cite{paull2013sensor,chen2020active,mansouri2018visual,rios2025zoom,papaioannou2023integrated}.
Maritime search formulations additionally address area allocation,
vehicle routing,
and cooperative target localization
\cite{ai2021coverage,li2023maritime,esmailifar2017cooperative}.
These methods predominantly optimize vehicle trajectories or treat the gimbal as a visibility constraint.
Recent visual-servoing studies improve quadrotor tracking from robocentric
or perception-aware formulations
\cite{li2023robocentric,qin2023perception},
but do not convert an arbitrary water polygon into a detectability-constrained scan
from a fixed coastal hover point.
AUSLUN instead treats pod pitch,
yaw,
and zoom as the search actions
and adapts each annular sweep to the intersection between the scan ring and the mission polygon.

\subsection{GNSS-Denied Localization and Aerial-Surface Cooperation}

GNSS-resilient marine navigation has used VIO,
marine-radar SLAM,
LiDAR-assisted fusion,
and vision-based docking
\cite{liu23vio,han2019coastal,shen2023lidar,volden2022vision}.
More generally,
range-based relative localization and multisource navigation fusion provide rigorous tools
for cooperative localization under uncertain measurements or GNSS interference
\cite{wang2023range,meng2025trusted}.
These approaches are complementary to,
but structurally different from,
the present problem:
the distant target is non-cooperative,
the USV is cooperative,
and the two objects therefore provide different range modalities.

UAV--USV studies have considered coordinated trajectory tracking,
aerial estimation of rescue-USV pose,
cooperative search and rescue,
and dynamics-aware joint planning
\cite{wang2021coordinated,dufek2019visual,wang2023cooperative,huang2023usv}.
Their primary focus is platform motion or cooperative-vehicle estimation.
AUSLUN uses the UAV as a fixed navigation anchor
and couples laser-assisted target localization,
datalink-assisted USV localization,
and target-relative command transmission in the same online sequence.

\subsection{Maritime Perception and Visual-Loss Handling}

Maritime UAV perception includes stabilized geolocation,
vision-range tracking,
onboard detection-tracking-localization,
prioritized vessel detection,
and floating-object tracking
\cite{wang2017multitarget,wang2018accurate,tang2020onboard,saoud2025prioritized,helgesen2021tracking}.
Visual servoing with explicit visibility constraints
\cite{xie2017input,zheng2019visibility}
and recent continuous image-based servoing under signal loss
\cite{he2026cibvs}
show the importance of preserving or recovering image measurements.
In AUSLUN,
recovery is not claimed as a standalone visual-servoing law.
It is a supervisory mechanism that distinguishes
between target loss during global search
and USV loss during guidance:
the former resumes the remaining search plan,
whereas the latter expands the FOV and performs a local spiral around the last valid direction.

In summary,
the paper's novelty is the constrained system architecture and its field-verified coupling:
fixed-hover pod coverage,
cooperation-dependent range sensing,
measurement-gated estimation,
relative guidance,
and task-specific loss transitions.

\section{System Configuration and Measurement Model}
\label{sec:scenario}

Building on the scenario introduced in Section~\ref{sec:introduction},
this section defines the coordinate frames,
platform configuration,
and sensing and communication models used in the problem formulation.

Three coordinate frames are used throughout this paper.
The local navigation frame $\mathcal{I}$ is a north-east-up frame
with origin at the UAV takeoff point and the water surface represented by $z=0$.
The UAV body frame $\mathcal{B}$ is a forward-left-up frame attached to the UAV,
and the camera frame $\mathcal{C}$ is a forward-left-up frame centered at the camera.
Let $\boldsymbol{p}_u = [x_u, y_u, h_u]^T$ denote the UAV position in $\mathcal{I}$,
where $h_u>0$ is the hover altitude.
Let $R_B^I$ denote the rotation from body frame to inertial frame,
and let $R_C^B$ denote the rotation from camera frame to body frame.
These matrices are constructed from the UAV attitude
and the pod angles.

The system contains two cooperative platforms and one stationary non-cooperative target.
The UAV carries a downward-facing VIO module,
a two-axis gimbaled pod,
a laser ranging module,
an onboard computer,
and a datalink.
The pod supports independent pitch and yaw control
and continuous optical zoom with horizontal field of view (HFOV) from \SI{2.36}{\degree}
to \SI{63.49}{\degree}.
Let $\mathcal{S}_c$ denote the set of coastal hover positions
where visual features are visible to the VIO system,
and let $h_{\mathrm{VIO}}$ denote the maximum altitude
for reliable low-altitude VIO.
The UAV position satisfies
\begin{equation}
\label{eq:vio_hover_constraint}
\begin{aligned}
\boldsymbol{p}_u&=[x_u,y_u,h_u]^T,\\
\text{s.t.}\quad
(x_u,y_u)&\in \mathcal{S}_c,\quad
h_u\in[h_{\min},h_{\mathrm{VIO}}].
\end{aligned}
\end{equation}
In the field implementation,
the UAV operates from a low-altitude fixed hover
where coastal features remain visible to the VIO system
and $\boldsymbol{p}_u$ anchors the bearing-range conversion
and target-relative navigation commands \cite{shan_lvi-sam_2021, weinstein_vio_2018}.

The USV is a cooperative platform.
It receives navigation commands from the UAV
and carries a datalink transponder
that provides range measurements over a nominal \SI{4}{\kilo\meter} link.
This provides the practical basis for using datalink range
during long-range USV navigation assistance.
In the GNSS-denied phase,
the USV relies on the UAV-estimated position and relative navigation vector.
The target vessel is stationary and non-cooperative.
It does not provide communication,
range,
or pose information.
Its position must be estimated from image-based bearing measurements
and laser ranging measurements collected after the pod is aligned with the vessel.

Two bearing-range measurement modes are used.
For the non-cooperative target vessel,
the UAV forms the measurement
\begin{equation}
\label{eq:target_measurement}
\boldsymbol{z}_t =
[\rho_l, \alpha_t, \epsilon_t, h_u]^T,
\end{equation}
where $\rho_l$ is the laser ranging measurement,
$\alpha_t$ is the camera azimuth (positive to the left)
and $\epsilon_t$ is the camera elevation angle (positive downward),
and $h_u$ is the UAV operating altitude.
The range component is used only after the pod has been aligned with the target
and a valid laser ranging measurement is available.
This alignment introduces a short dwell period,
but it is necessary because the target vessel does not cooperate with the localization system.

For the cooperative USV,
the UAV forms the measurement
\begin{equation}
\label{eq:usv_measurement}
\boldsymbol{z}_v =
[\rho_v, \alpha_v, \epsilon_v, h_u]^T,
\end{equation}
where $\rho_v$ is the datalink range to the USV.
The same bearing-range geometry is used for both vessels,
while the range input depends on whether the vessel is cooperative.

\section{Problem Formulation}
\label{sec:problem}

Based on the system configuration in Section~\ref{sec:scenario},
the objective is to design an online fixed-hover UAV policy
for target search,
target and USV localization,
and USV navigation assistance
in a system with a cooperative USV
and a non-cooperative target vessel.
Given the search region $\Omega$,
the VIO-feasible fixed UAV position $\boldsymbol{p}_u$,
the pod actuation limits,
a target image template,
the USV dock region,
and the available laser and datalink range measurements,
the policy must generate pod commands,
form vessel measurements,
estimate vessel positions,
and publish USV navigation commands
until the USV reaches the close-range acquisition region.
The control command of the pod is
\begin{equation}
\label{eq:pod_command}
\boldsymbol{u}_p(t) =
[\theta_p(t), \theta_y(t), \phi_h(t)]^T,
\end{equation}
where $\theta_p$ is pitch (positive downward),
$\theta_y$ is yaw (positive to the left),
and $\phi_h$ is horizontal FOV.
The coupled policy is decomposed below into search,
target localization,
USV localization and navigation,
and visual-loss recovery subproblems.

\subsection{Pod-Based Search Problem}

Given the search region $\Omega$,
the VIO-feasible fixed UAV position $\boldsymbol{p}_u$,
and the admissible pod command set $\mathcal{A}_p$
defined by the pod pitch,
yaw,
and FOV limits,
the search problem is to generate a finite pod command sequence
\begin{equation}
\label{eq:scan_sequence}
\mathcal{U} =
\{\boldsymbol{u}_p^1, \boldsymbol{u}_p^2, \ldots, \boldsymbol{u}_p^N\}
\end{equation}
that covers $\Omega$
while avoiding unnecessary scans outside the polygonal search region.
Let $\mathcal{F}_i=\mathcal{F}(\boldsymbol{u}_p^i;\boldsymbol{p}_u)$.
The coverage requirement is expressed as
\begin{equation}
\label{eq:coverage_requirement}
\begin{aligned}
&\Omega \subseteq
\bigcup_{i=1}^{N}\mathcal{F}_i,\\
&
\boldsymbol{u}_p^i \in \mathcal{A}_p,\quad
i=1,\ldots,N,
\end{aligned}
\end{equation}
where $\mathcal{F}(\cdot)$ is the camera footprint on the water plane.
The multi-ring annular construction and overlap rule
are method choices introduced in Section~\ref{sec:method}.

\subsection{Non-Cooperative Target Localization Problem}

After a candidate target vessel is detected,
the system must estimate its position
$\boldsymbol{p}_t \in \mathbb{R}^3$
from the target measurement $\boldsymbol{z}_t$ in \eqref{eq:target_measurement}.
The range component is obtained only after visual tracking,
pod alignment,
and a returned laser ranging measurement.
Let $c_t$ denote the visual confidence,
$(u_t,v_t)$ the normalized target point in the image
($u_t$ positive to the right, $v_t$ positive downward),
$\tau_c$ the minimum visual-confidence threshold,
$\tau_a$ the normalized image-center alignment tolerance,
and $\chi_p$,
$\chi_y$,
and $\chi_\phi$ denote whether the pitch,
yaw,
and FOV commands have reached their targets.
Let $a_l(t)\in\{0,1\}$ indicate whether a laser ranging measurement is returned
for the currently tracked vessel.
The target measurement condition is
\begin{equation}
\label{eq:target_measurement_condition}
\begin{aligned}
\mathcal{C}_t(t):\quad
&c_t \geq \tau_c,\quad
\sqrt{u_t^2+v_t^2}\leq \tau_a,\\
&\chi_p\chi_y\chi_\phi=1,\quad
a_l(t)=1.
\end{aligned}
\end{equation}
This condition captures the implemented dwell-and-ranging procedure:
the pod first tracks the visual target,
then waits until the pod attitude and zoom commands are reached,
and only then uses the laser ranging measurement for target localization.
The target localization problem is to estimate
\begin{equation}
\label{eq:target_estimate}
\hat{\boldsymbol{p}}_t =
f_t(\boldsymbol{z}_t, \boldsymbol{p}_u, R_B^I, R_C^B),
\end{equation}
using only frames that contain a laser ranging measurement
and sufficient visual confidence.

\subsection{USV Localization and Navigation Problem}

The cooperative USV position is denoted by $\boldsymbol{p}_v \in \mathbb{R}^3$.
Given the measurement $\boldsymbol{z}_v$ in \eqref{eq:usv_measurement},
the UAV estimates the USV position as
\begin{equation}
\label{eq:usv_estimate}
\hat{\boldsymbol{p}}_v =
f_v(\boldsymbol{z}_v, \boldsymbol{p}_u, R_B^I, R_C^B).
\end{equation}
Let $c_v$ denote the visual confidence for the cooperative USV,
let $(u_v,v_v)$ denote the normalized USV tracking point,
and let $a_d(t)\in\{0,1\}$ indicate whether a datalink range is received.
The USV measurement condition is
\begin{equation}
\label{eq:usv_measurement_condition}
\begin{aligned}
\mathcal{C}_v(t):\quad
&c_v \geq \tau_c,\quad
\sqrt{u_v^2+v_v^2}\leq \tau_a,\\
&a_d(t)=1 .
\end{aligned}
\end{equation}
The same confidence and alignment thresholds
are used as in \eqref{eq:target_measurement_condition}.
Unlike the non-cooperative target condition in \eqref{eq:target_measurement_condition},
\eqref{eq:usv_measurement_condition} pairs the visual bearing with the datalink range
for cooperative-USV localization and navigation.
Once $\hat{\boldsymbol{p}}_t$ and $\hat{\boldsymbol{p}}_v$ are available,
the relative navigation vector is
\begin{equation}
\label{eq:navigation_vector}
\boldsymbol{n}_{v \rightarrow t}
= [n_x, n_y, n_z]^T
= \hat{\boldsymbol{p}}_t - \hat{\boldsymbol{p}}_v,
\end{equation}
and the desired heading command is
\begin{equation}
\label{eq:navigation_heading}
\psi_n =
\operatorname{atan2}(n_y, n_x).
\end{equation}
The UAV sends $\boldsymbol{n}_{v \rightarrow t}$ and $\psi_n$
to the USV through the datalink.
The navigation stage is complete when the cooperative USV reports
a local LiDAR acquisition signal,
after which the USV takes over the final approach.

\subsection{Recovery Problem}

When visual tracking is lost,
the system must reacquire the target vessel or the USV
without restarting the full search over $\Omega$.
Given the last valid pod command $\boldsymbol{u}_{p,0}$
and the last known visual target identity,
the recovery problem is to generate a local scan command
that searches the neighborhood of the last visual direction
without restarting a full-area search.

\subsection{System-Level Problem Summary}

The complete policy must therefore generate a scan sequence $\mathcal{U}$,
estimate the target and USV through $f_t$ and $f_v$,
publish target-relative guidance,
and select the appropriate recovery transition.
Throughout the mission,
the UAV must remain inside the VIO-feasible hover set,
the planned footprints must cover $\Omega$ under the adopted footprint model,
and estimator updates are permitted only when $\mathcal{C}_t$ or $\mathcal{C}_v$ is satisfied.
The mission terminates when the USV reports close-range LiDAR acquisition.

\section{Proposed Method}
\label{sec:method}

\begin{figure*}[!t]
\centering
\includegraphics[width=\textwidth]{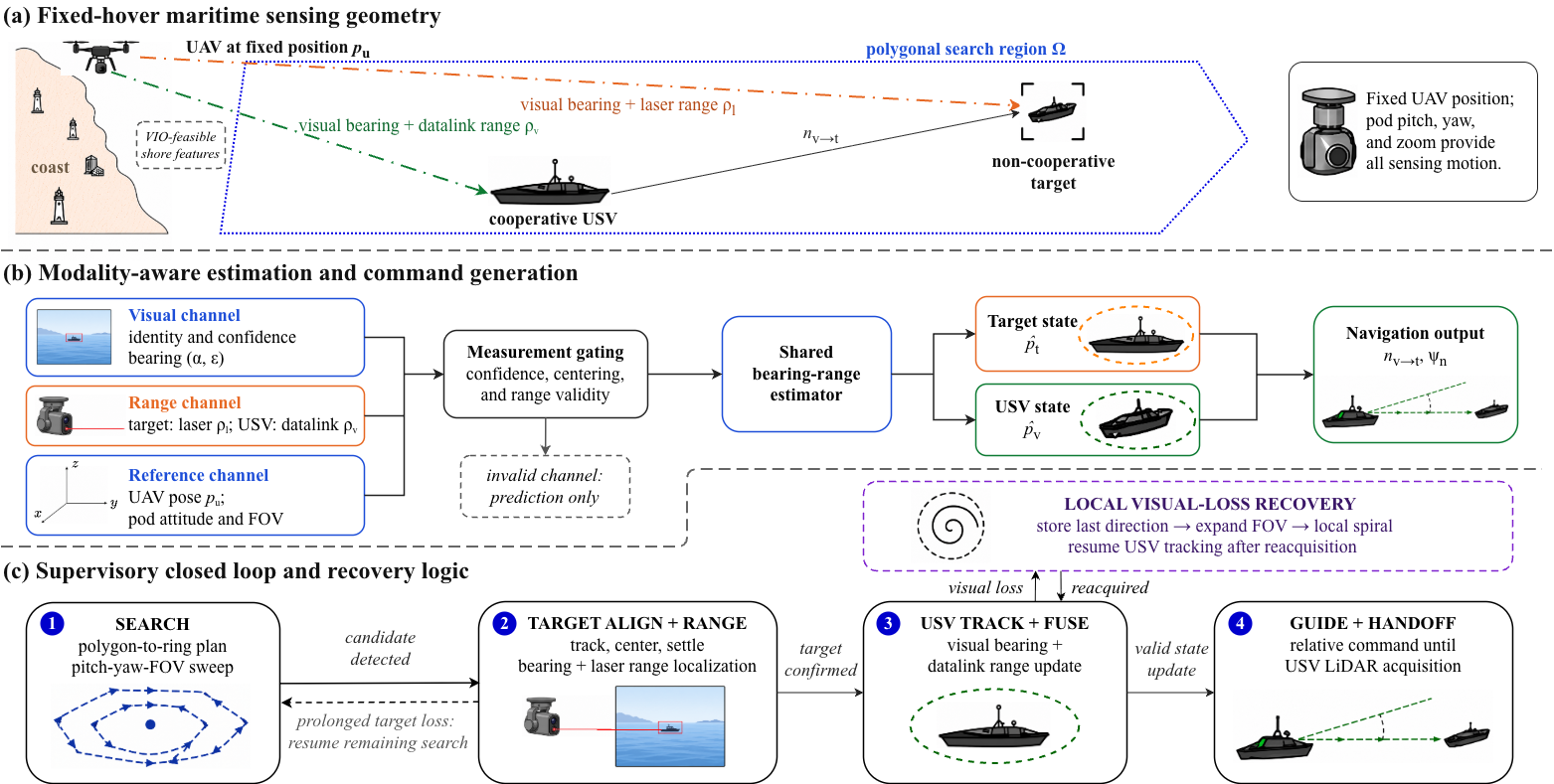}
\caption{Closed-loop architecture of AUSLUN.
(a) A coastal fixed-hover UAV transfers sensing motion to the pitch-yaw-zoom pod
and uses different range channels for the non-cooperative target and cooperative USV.
(b) Measurement gating feeds a common bearing-range recursive estimator,
whose target and USV outputs generate relative guidance.
(c) The supervisory state machine separates global target search,
laser-gated target localization,
datalink-assisted USV tracking,
local visual-loss recovery,
and close-range LiDAR handoff.}
\label{fig:task_flow}
\end{figure*}

Fig.~\ref{fig:task_flow} organizes AUSLUN as a closed perception-estimation-guidance loop.
The fixed UAV pose and pod geometry define the search action space;
visual,
laser,
and datalink channels enter a modality-aware estimator;
and a supervisory state machine determines whether the system searches,
localizes,
guides,
recovers,
or hands control to onboard USV sensing.
The method receives the search polygon,
fixed UAV pose,
pod limits,
target template,
USV dock region,
and asynchronous measurements.
It returns the target and USV state estimates
and the corresponding target-relative USV command.

\subsection{FOV-Aware Pod Search}

Because the UAV position is fixed by the VIO constraint,
the search planner generates pod commands rather than UAV waypoints.
It models the camera footprint as a function of UAV altitude,
camera pitch,
and FOV.
For a given camera pitch $\theta_p$,
the distance from the UAV to the ground footprint center is
\begin{equation}
\label{eq:range_from_pitch}
r(\theta_p) = \frac{h_u}{\tan(\theta_p)}.
\end{equation}
The horizontal extent of the footprint at this distance is
\begin{equation}
\label{eq:horizontal_footprint}
L_h(\theta_p) =
2 r(\theta_p)
\tan\left(\frac{\phi_h(\theta_p)}{2}\right),
\end{equation}
and the vertical extent is
\begin{equation}
\label{eq:vertical_footprint}
L_v(\theta_p) =
2 r(\theta_p)
\tan\left(\frac{\phi_v}{2}\right).
\end{equation}
Here $\phi_v$ is obtained from the horizontal FOV and the image aspect ratio.
The water surface is approximated as locally planar,
and these footprint dimensions are planning approximations for overlap-controlled scan rings;
$\theta_p$ is the positive downward pitch angle from the horizontal line of sight.
The planner uses three search-design parameters
to connect geometric coverage with visual detectability.
The apparent-size parameter $\gamma_w$ specifies
the minimum fraction of the image width
that a vessel with nominal length $L_\mathrm{tar}$ should occupy.
Accordingly,
the horizontal footprint is selected to satisfy
\begin{equation}
\label{eq:detection_constraint}
\frac{L_\mathrm{tar}}{L_h(\theta_p)}
\geq
\gamma_w,
\quad
L_h(\theta_p)
\leq
\frac{L_\mathrm{tar}}{\gamma_w}.
\end{equation}
The planner chooses the widest admissible FOV
that satisfies \eqref{eq:detection_constraint}
within the pod FOV limits.
The sweep-time parameter $T_s$ specifies
how long a target should remain inside one horizontal FOV
during a yaw sweep.
The overlap parameter $\eta$ specifies
the required vertical overlap between adjacent pitch rings,
so that multi-ring scanning does not leave uncovered strips
between neighboring rings.

To instantiate the generic coverage problem in Section~\ref{sec:problem},
the search area is divided into annular regions centered at the UAV position.
The rings are ordered from far to near,
\begin{equation}
\label{eq:far_to_near_order}
r_1 \geq r_2 \geq \cdots \geq r_N,
\end{equation}
so that the pod first scans the farthest visible annulus
and then moves inward.
For the $i$-th ring,
the pitch angle is computed from its center range $r_i$:
\begin{equation}
\label{eq:pitch_sequence}
\theta_p^{(i)} =
\arctan\left(\frac{h_u}{r_i}\right).
\end{equation}
Adjacent rings satisfy the overlap constraint
\begin{equation}
\label{eq:overlap_constraint}
\begin{aligned}
r_i - r_{i+1}
&\leq
\frac{1-\eta}{2}
\left(
L_v(\theta_p^{(i)})
+
L_v(\theta_p^{(i+1)})
\right),\\
&\quad i=1,\ldots,N-1.
\end{aligned}
\end{equation}
For each ring,
the yaw bounds are obtained by intersecting the search polygon
with the circular arc at range $r_i$:
\begin{equation}
\label{eq:yaw_range}
\theta_y \in
[\theta_{y,\min}^{(i)}, \theta_{y,\max}^{(i)}].
\end{equation}
The horizontal yaw rate for the $i$-th ring is
\begin{equation}
\label{eq:yaw_sweep_rate}
\omega_y^{(i)}
=
\frac{\phi_h(\theta_p^{(i)})}{T_s},
\end{equation}
and the resulting search time is approximated by
\begin{equation}
\label{eq:total_time}
T_{\mathrm{search}}
=
\sum_{i=1}^{N}
\frac{|\theta_{y,\max}^{(i)}-\theta_{y,\min}^{(i)}|}
{\phi_h(\theta_p^{(i)})}
T_s,
\end{equation}
excluding the short repositioning commands
used to move the pod to the start of each ring.

\subsection{Vision-Ranging Localization}

After a candidate target vessel is detected,
the pod enters a tracking-and-ranging mode.
The visual detector provides a target identity,
a confidence or template-matching score,
and a bounding box.
The controller selects a tracking point inside the box
and keeps it near the image center before using range measurements.
Let $(u_t, v_t) \in [-1,1]^2$ denote the normalized tracking point
in the camera image.
The target bearing angles are computed as
\begin{equation}
\label{eq:target_pixel_to_angles}
\begin{cases}
\alpha_t =
-\arctan\left(u_t \tan\dfrac{\phi_h}{2}\right),\\[6pt]
\epsilon_t =
\arctan\left(v_t \tan\dfrac{\phi_v}{2}\right).
\end{cases}
\end{equation}
The pod controller updates its pitch,
yaw,
and zoom commands until the target is close to the image center.
In the implemented controller,
the desired pod command is
\begin{equation}
\label{eq:pod_alignment_command}
\theta_p^\star = \theta_p+\epsilon_t,
\qquad
\theta_y^\star = \theta_y+\alpha_t,
\end{equation}
and the zoom is adjusted according to the target-box width
so that the target remains within a useful image scale.
When the box width falls below the lower bound in Table~\ref{tab:simulation_parameters},
the controller reduces the horizontal FOV;
when it exceeds the upper bound,
the controller increases the horizontal FOV.
The alignment flags from the pod controller are used as a dwell condition,
so that laser ranging is used only after the optical axis
and the target point are sufficiently aligned.
In the field implementation,
this centering requirement is enforced by the tracking command
and pod-settling flags,
rather than by a separate hard-coded image-coordinate threshold.
Laser ranging is then enabled,
and the frame is used for localization only when a laser ranging measurement is obtained
and the target remains visible.
Missing laser ranging measurements,
missing detections,
or incomplete pod alignment cause the frame to be treated as prediction-only
in the estimator rather than as a localization update.
If visual tracking of the target vessel is lost for a timeout period,
the controller falls back to the remaining search sequence
instead of accepting a bearing-only target position.
The target position in the camera frame is
\begin{equation}
\label{eq:target_camera_position}
\boldsymbol{p}_{t}^{C}
=
\begin{bmatrix}
1\\
\tan\alpha_t\\
-\tan\epsilon_t
\end{bmatrix}
\frac{\rho_l}
{\sqrt{1+\tan^2\alpha_t+\tan^2\epsilon_t}}.
\end{equation}
This expression converts the measured range and bearing angles
into a camera-frame position vector.
The corresponding inertial position is
\begin{equation}
\label{eq:target_inertial_position}
\hat{\boldsymbol{p}}_t
=
\boldsymbol{p}_u + R_B^I R_C^B \boldsymbol{p}_{t}^{C}.
\end{equation}
The estimator below smooths this bearing-range estimate
when multiple measurement frames are available.

The same bearing-range form is used for both target and USV localization.
For a tracked object $j \in \{t,v\}$,
the measurement is
\begin{equation}
\label{eq:measurements}
\boldsymbol{z}_j =
[\rho_j, \alpha_j, \epsilon_j, h_u]^T.
\end{equation}
For the target vessel,
$\rho_j=\rho_l$.
For the USV,
$\rho_j=\rho_v$ is the datalink range.
The datalink value is used as a quantized cooperative range input,
not as a high-resolution ranging sensor;
$\boldsymbol{R}=\operatorname{diag}(\sigma_{\rho}^{2},\sigma_{\alpha}^{2},
\sigma_{\epsilon}^{2},\sigma_{h}^{2})$
weights the range, azimuth, elevation, and altitude measurement noise.

The estimator state is
\begin{equation}
\label{eq:state_vector}
\boldsymbol{x}_j =
[x_j, y_j, z_j, v_{x,j}, v_{y,j}, v_{z,j}]^T.
\end{equation}
A nearly constant velocity model is used:
\begin{equation}
\label{eq:process_model}
\boldsymbol{x}_{j,k+1}
=
\boldsymbol{F}_k \boldsymbol{x}_{j,k}
+ \boldsymbol{w}_k,
\end{equation}
where $\boldsymbol{F}_k$ is the discrete constant-velocity transition matrix
and $\boldsymbol{w}_k \sim \mathcal{N}(\boldsymbol{0},\boldsymbol{Q})$.
The nonlinear measurement model is
\begin{equation}
\label{eq:measurement_function}
\boldsymbol{g}(\boldsymbol{x}_j)
=
\begin{bmatrix}
\sqrt{x_c^2+y_c^2+z_c^2}\\[4pt]
\operatorname{atan2}(y_c,x_c)\\[4pt]
-\operatorname{atan2}(z_c,x_c)\\[4pt]
h_u
\end{bmatrix},
\end{equation}
where $[x_c,y_c,z_c]^T=\boldsymbol{p}_{j}^{C}$.
The fourth component reduces to the known altitude $h_u$
under the planar-water assumption $z_j\approx 0$.
The camera-frame position is obtained from the inertial-frame state by
\begin{equation}
\label{eq:camera_coordinate}
\boldsymbol{p}_{j}^{C}
=
R_B^C R_I^B
(\boldsymbol{x}_{j,1:3}-\boldsymbol{p}_u).
\end{equation}
Here $R_B^C=(R_C^B)^T$ and $R_I^B=(R_B^I)^T$.
The Jacobian $\boldsymbol{H}_k =
\partial \boldsymbol{g}/\partial \boldsymbol{x}_j|_{\bar{\boldsymbol{x}}_{j,k}}$
linearizes this measurement model around the predicted state.
The recursive prediction and update are
\begin{equation}
\label{eq:estimator_prediction}
\begin{aligned}
\bar{\boldsymbol{x}}_{j,k}
&= \boldsymbol{F}_k \boldsymbol{x}_{j,k-1},\\
\bar{\boldsymbol{P}}_{j,k}
&= \boldsymbol{F}_k \boldsymbol{P}_{j,k-1} \boldsymbol{F}_k^T
+ \boldsymbol{Q},
\end{aligned}
\end{equation}
and
\begin{equation}
\label{eq:estimator_update}
\begin{aligned}
\boldsymbol{K}_k
&=
\bar{\boldsymbol{P}}_{j,k}
\boldsymbol{H}_k^T
(\boldsymbol{H}_k \bar{\boldsymbol{P}}_{j,k}\boldsymbol{H}_k^T
+\boldsymbol{R})^{-1},\\
\boldsymbol{x}_{j,k}
&=
\bar{\boldsymbol{x}}_{j,k}
+ \boldsymbol{K}_k
(\boldsymbol{z}_{j,k}-\boldsymbol{g}(\bar{\boldsymbol{x}}_{j,k})),\\
\boldsymbol{P}_{j,k}
&=
(\boldsymbol{I}-\boldsymbol{K}_k \boldsymbol{H}_k)
\bar{\boldsymbol{P}}_{j,k}.
\end{aligned}
\end{equation}

\subsection{Datalink-Assisted Navigation}

After the target position and USV position are estimated,
the UAV computes the relative navigation vector using \eqref{eq:navigation_vector}.
The heading command in \eqref{eq:navigation_heading}
is transmitted to the USV together with the relative position.
This separates non-cooperative target localization
from cooperative USV navigation while retaining a common bearing-range model.
In navigation mode,
the datalink range is used with the current visual bearing.
This preserves the cooperative USV localization chain
without introducing a second USV ranging mode.

\subsection{Visual-Loss Recovery}

Visual tracking loss is detected when no valid detection is received
for a timeout period $T_{\mathrm{loss}}$
or the detection confidence falls below the threshold $\tau_c$.
The system stores the last valid pod command
$(\theta_{p,0}, \theta_{y,0}, \phi_{h,0})$
and expands the horizontal FOV as
\begin{equation}
\label{eq:fov_expansion}
\phi_h'
=
\min(\phi_h^{\max}, \kappa \phi_{h,0}),
\end{equation}
where $\kappa$ is the FOV expansion factor.
If the target is not recovered through FOV expansion,
the pod performs a sinusoidal spiral scan:
\begin{equation}
\label{eq:spiral_scan}
\begin{cases}
\theta_p(t) =
\theta_{p,0}
+ \dot{A}_p t
\sin\left(\dfrac{2\pi t}{T_{\mathrm{scan}}}\right),\\[8pt]
\theta_y(t) =
\theta_{y,0}
+ \dot{A}_y t
\cos\left(\dfrac{2\pi t}{T_{\mathrm{scan}}}\right).
\end{cases}
\end{equation}
Here $\dot{A}_p$ and $\dot{A}_y$ are pitch and yaw envelope expansion rates.
The scan expands around the last known visual direction
and returns to the previous tracking or navigation state
after the target vessel or USV is reacquired.
For the target-localization stage,
a prolonged loss of visual detections causes the controller
to resume the remaining search sequence.
For the USV-navigation stage,
the controller enters the local visual-loss recovery state
because the USV identity and approximate direction are already known.

Algorithm~\ref{alg:framework} summarizes the complete online workflow.
\begin{algorithm}[!t]
\caption{UAV-Assisted Search, Target Localization, and USV Navigation}
\label{alg:framework}
\begin{algorithmic}
\STATE \emph{Input:} $\Omega$, $\boldsymbol{p}_u$, FOV limits, target template, USV dock region
\STATE \emph{Output:} $\hat{\boldsymbol{p}}_t$, $\hat{\boldsymbol{p}}_v$, $\boldsymbol{n}_{v \rightarrow t}$, $\psi_n$
\STATE Generate FOV-aware pod scan sequence $\mathcal{U}$
\FOR{each command $\boldsymbol{u}_p^i \in \mathcal{U}$}
    \STATE Execute pod search command $\boldsymbol{u}_p^i$
    \IF{candidate target vessel is detected}
        \STATE Track the target and align the pod with the target point
        \STATE Enable laser ranging and use $\boldsymbol{z}_t$ only when $\mathcal{C}_t$ holds
        \STATE Estimate $\hat{\boldsymbol{p}}_t$ from $\boldsymbol{z}_t$
        \IF{target is confirmed}
            \STATE Reorient the pod toward the USV dock region
            \WHILE{USV local LiDAR acquisition signal is not received}
                \STATE Track the USV and use $\boldsymbol{z}_v$ when $\mathcal{C}_v$ holds
                \STATE Estimate $\hat{\boldsymbol{p}}_v$
                \STATE Compute $\boldsymbol{n}_{v \rightarrow t}$ and $\psi_n$
                \STATE Publish the navigation command to the USV
                \IF{visual tracking is lost}
                    \STATE Execute FOV expansion and sinusoidal spiral recovery
                \ENDIF
            \ENDWHILE
            \STATE \textsc{break}
        \ENDIF
    \ENDIF
\ENDFOR
\end{algorithmic}
\end{algorithm}

\section{Experimental Evaluation}
\label{sec:experiments}

The evaluation follows the paper's claim hierarchy.
First,
deterministic planning simulations test whether polygon-dependent yaw bounds
reduce scan time and redundant coverage under matched sensing constraints.
Second,
an integrated Yas Island mission tests whether the complete state sequence
can progress from search to target localization,
USV guidance,
recovery,
and close-range handoff.
Third,
a deliberately activated interruption tests the recovery-to-navigation transition.
Finally,
a GPS-referenced field segment tests localization accuracy against non-recursive baselines.
Because the available field record contains one integrated mission
and one deliberately activated recovery interval,
the corresponding results establish feasibility and observed behavior,
not a statistical reliability rate.

\subsection{Search-Planning Simulation}

\subsubsection{Setup}

The search-planning simulation uses the system configuration in
Section~\ref{sec:scenario}
and the search problem defined in Section~\ref{sec:problem}.
The field-polygon case uses the same polygonal search region
as the Yas Island deployment
and the same design and implementation parameters as the field system,
summarized in Table~\ref{tab:simulation_parameters};
the baseline is a fixed-sector angular raster scan with matched pitch rings
and dwell time.

\begin{table}[!t]
\caption{Design and implementation parameters.
\label{tab:simulation_parameters}}
\centering
\scriptsize
\setlength{\tabcolsep}{2pt}
\begin{tabular}{@{}
>{\raggedright\arraybackslash}p{0.30\columnwidth}
>{\centering\arraybackslash}p{0.22\columnwidth}
>{\raggedright\arraybackslash}p{0.40\columnwidth}@{}}
\toprule
Parameter & Symbol & Value \\
\midrule
Search polygon vertices
& $n_\Omega$
& $4$ \\
Search polygon area
& $|\Omega|$
& \SI{2.45}{\kilo\meter\squared} \\
Search bounding box
& $b_\Omega$
& \SI{1.53}{\kilo\meter}$\times$\SI{1.78}{\kilo\meter} \\
Fixed-hover UAV position
& $\boldsymbol{p}_u$
& $(0,0,7.5)\,\si{\meter}$ \\
Search yaw span
& $\Delta\theta_y$
& \SI{180}{\degree} \\
Image aspect ratio
& $a_{\mathrm{img}}$
& $16:9$ \\
Nominal target length
& $L_{\mathrm{tar}}$
& \SI{10}{\meter} \\
Minimum target width fraction
& $\gamma_w$
& $0.1$ \\
Pitch-ring overlap
& $\eta$
& $0.1$ \\
In-view time
& $T_s$
& \SI{4}{\second} \\
Target box width fraction
& $w_t$
& $0.10$--$0.50$ \\
USV box width fraction
& $w_v$
& $0.08$--$0.12$ \\
Visual-confidence threshold
& $\tau_c$
& $0.9$ \\
Image-center alignment tolerance
& $\tau_a$
& $0.05$ \\
Range update rate
& $f_{\rho_l},f_{\rho_v}$
& \SI{1}{\hertz} \\
Datalink range quantization
& $q_{\rho_v}$
& \SI{10}{\meter} \\
Measurement std.\ dev.
& $\sigma_{\rho},\sigma_{\alpha},\sigma_{\epsilon},\sigma_{h}$
& (\SI{1.0}{\meter}, \SI{0.71}{\degree}, \SI{0.71}{\degree}, \SI{2.24}{\meter}) \\
Process acceleration
& $\sigma_a$
& \SI{1.0}{\meter\per\second\squared} \\
Target visual-loss timeout
& $T_{\mathrm{loss},t}$
& \SI{20}{\second} \\
USV visual-loss timeout
& $T_{\mathrm{loss},v}$
& \SI{5}{\second} \\
FOV expansion factor
& $\kappa$
& $1.5$ \\
Recovery scan period
& $T_{\mathrm{scan}}$
& \SI{15}{\second} \\
Pitch envelope rate
& $\dot{A}_p$
& \SI{0.01}{\degree\per\second} \\
Yaw envelope rate
& $\dot{A}_y$
& \SI{0.033}{\degree\per\second} \\
Recovery rate limit
& $\dot{\theta}_{\max}$
& \SI{10}{\degree\per\second} \\
\bottomrule
\end{tabular}
\end{table}

The evaluation metrics are search time,
coverage completeness,
and redundant coverage.
The coverage planner is compared with a fixed-sector angular raster scan.
Localization and recovery are evaluated with field data in the following subsections.

\subsubsection{Results}

\begin{figure*}[!t]
\centering
\subfloat[Yas Island field polygon.]{
    \includegraphics[width=0.32\textwidth]{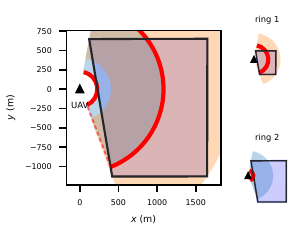}}
\hfill
\subfloat[Coastal corridor.]{
    \includegraphics[width=0.32\textwidth]{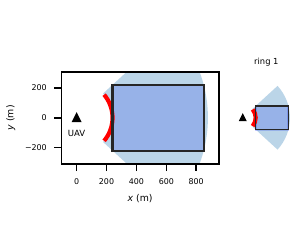}}
\hfill
\subfloat[Offshore bay.]{
    \includegraphics[width=0.32\textwidth]{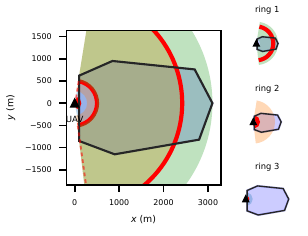}}
\caption{FOV-aware pod-search plans for three polygonal regions.
Each panel shows the combined scan plan and per-ring decomposition;
colored annular sectors are planned footprints,
and red arcs are active yaw sweeps.}
\label{fig:polygon_search_plans}
\end{figure*}

Fig.~\ref{fig:polygon_search_plans} shows three representative cases
generated from the same fixed-hover UAV position and search-design parameters.
The yaw bounds are computed from the intersection between each annular ring
and the polygonal mission region,
so the pod does not execute a full-sector scan when only a subset of the
sector belongs to the search area.

Table~\ref{tab:polygon_search_results} summarizes the planning results.
The three cases require one to three pitch rings depending on their radial span
from the fixed-hover UAV position.
Under the planner's footprint model,
all three cases cover the polygonal region
with the prescribed vertical-overlap requirement.
The fixed-sector angular raster baseline follows the conventional raster
or lawnmower idea \cite{galceran2013survey}
but uses the same fixed-hover viewpoint,
pitch rings,
zoom schedule,
overlap requirement,
and in-view time while sweeping each full yaw sector.
Compared with this baseline,
the adaptive yaw bounds reduce the planned scan time by
10.9\%--55.6\%
and reduce the redundant scanned area in all three regions.

\begin{table}[!t]
\caption{Search-planning results under matched pitch rings,
FOV schedule,
overlap,
and dwell time.
Scan time reports AUSLUN/fixed-sector baseline.
\label{tab:polygon_search_results}}
\centering
\scriptsize
\setlength{\tabcolsep}{2.2pt}
\begin{tabular}{@{}lccccc@{}}
\toprule
Case & Area & Rings & Scan time & Time red. & Area red. \\
& (\si{\kilo\meter\squared}) & & A/base (\si{\second}) & (\%) & (\%) \\
\midrule
Yas Island field polygon & 2.45 & 2 & 355.2/439.9 & 19.3 & 28.3 \\
Coastal corridor & 0.27 & 1 & 63.7/143.3 & 55.6 & 69.4 \\
Offshore bay & 5.12 & 3 & 591.0/663.5 & 10.9 & 13.3 \\
\bottomrule
\end{tabular}
\end{table}

\subsection{Field Experiments}

\subsubsection{Setup}

The field UAV is a DJI Matrice~300 equipped with a two-axis zoom pod,
a laser ranging module,
an onboard computer,
a datalink,
and the downward-facing VIO module described in Section~\ref{sec:scenario}.
The UAV runs the search,
tracking,
localization,
navigation,
and recovery modules onboard under Robot Operating System Noetic.
The cooperative USV is based on the Drone Carrier integrated USV platform
reported in \cite{dong2026dronecarrier};
in the present experiments,
it receives target-relative navigation commands from the UAV
and uses its onboard LiDAR for close-range acquisition.

\subsubsection{Integrated Field Mission}

\begin{figure*}[!t]
\centering
\subfloat[Search sweep.]{
    \includegraphics[width=0.235\textwidth]{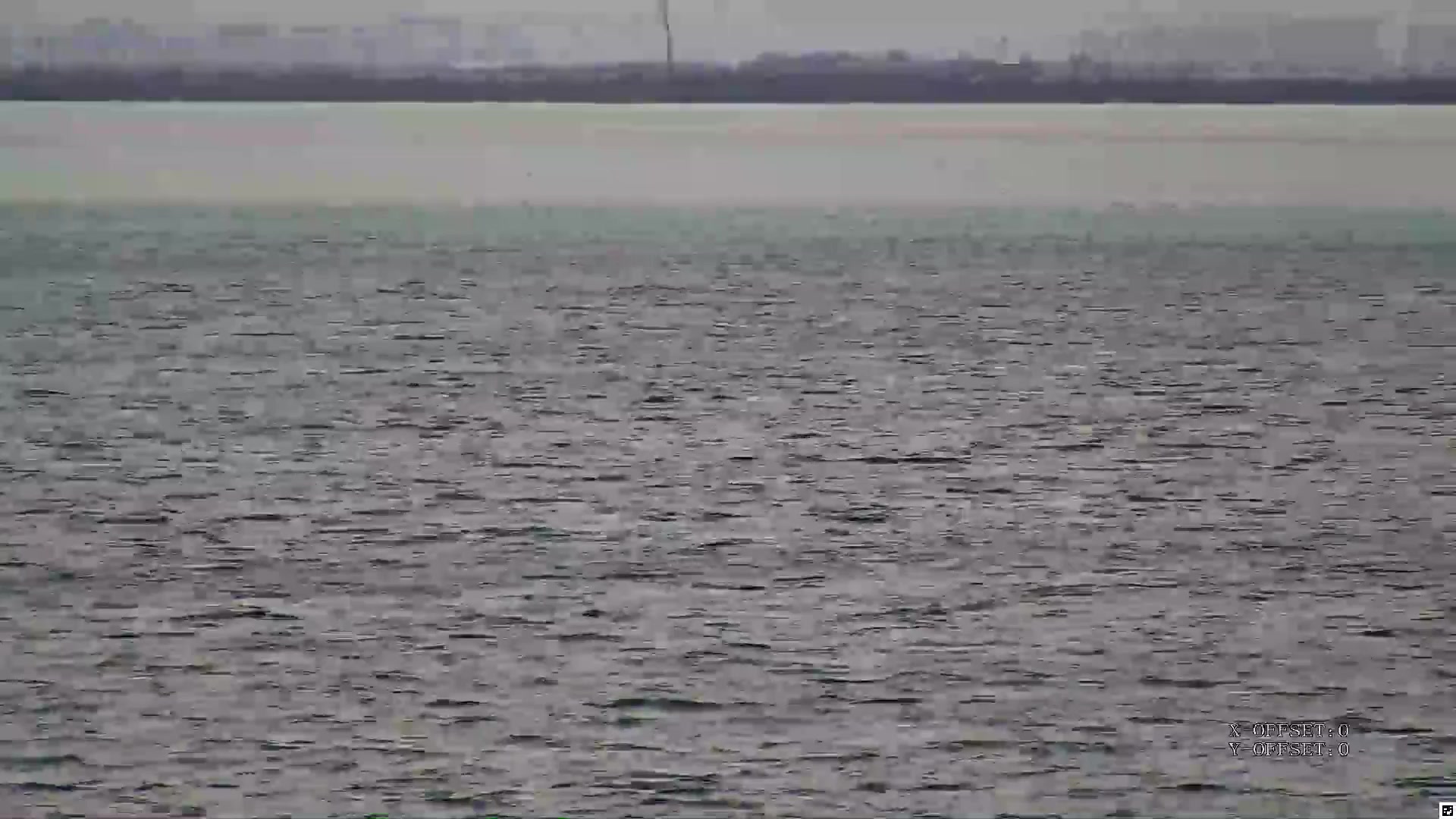}}
\hfill
\subfloat[Target entering.]{
    \includegraphics[width=0.235\textwidth]{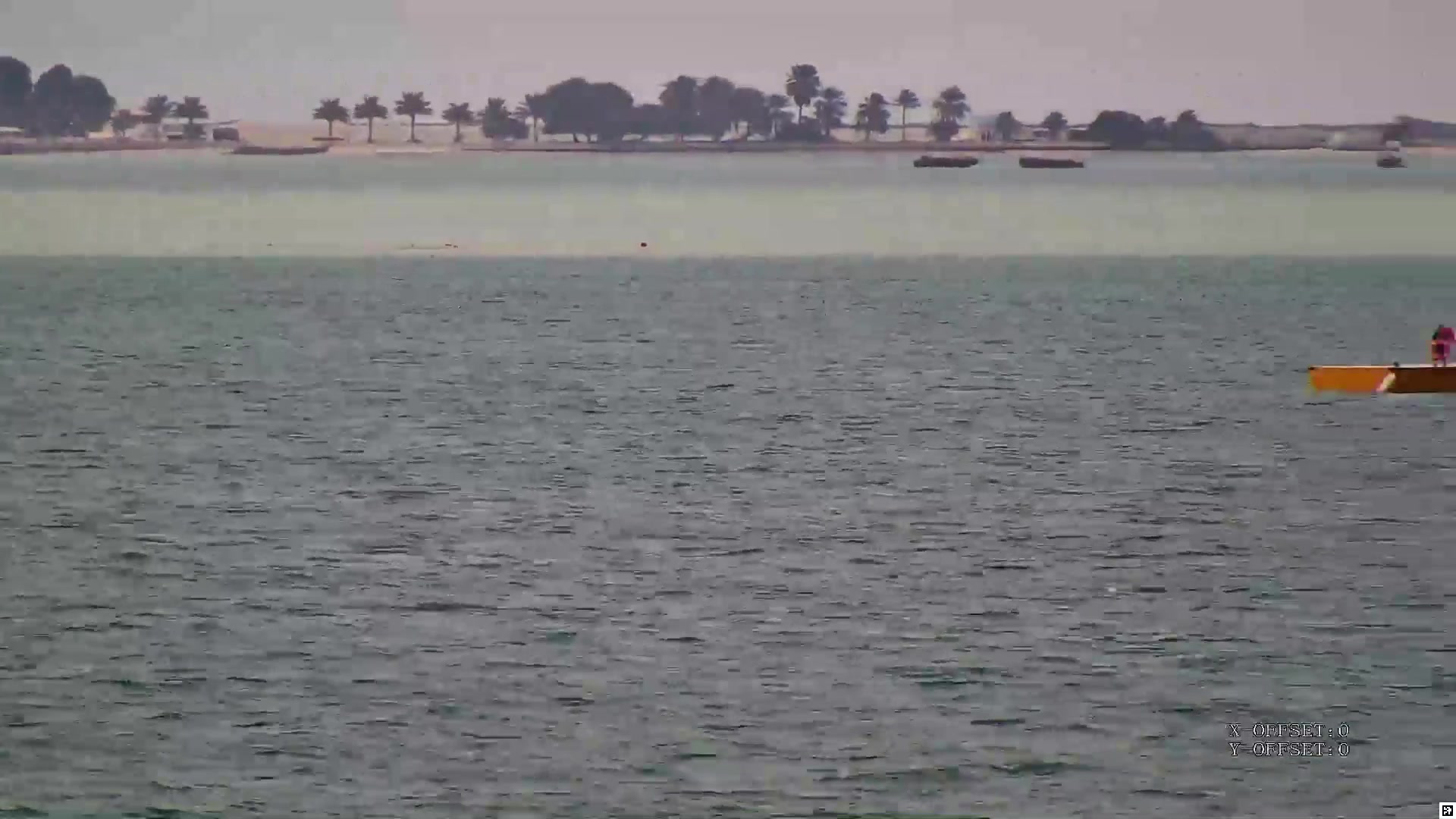}}
\hfill
\subfloat[Target confirmation.]{
    \includegraphics[width=0.235\textwidth]{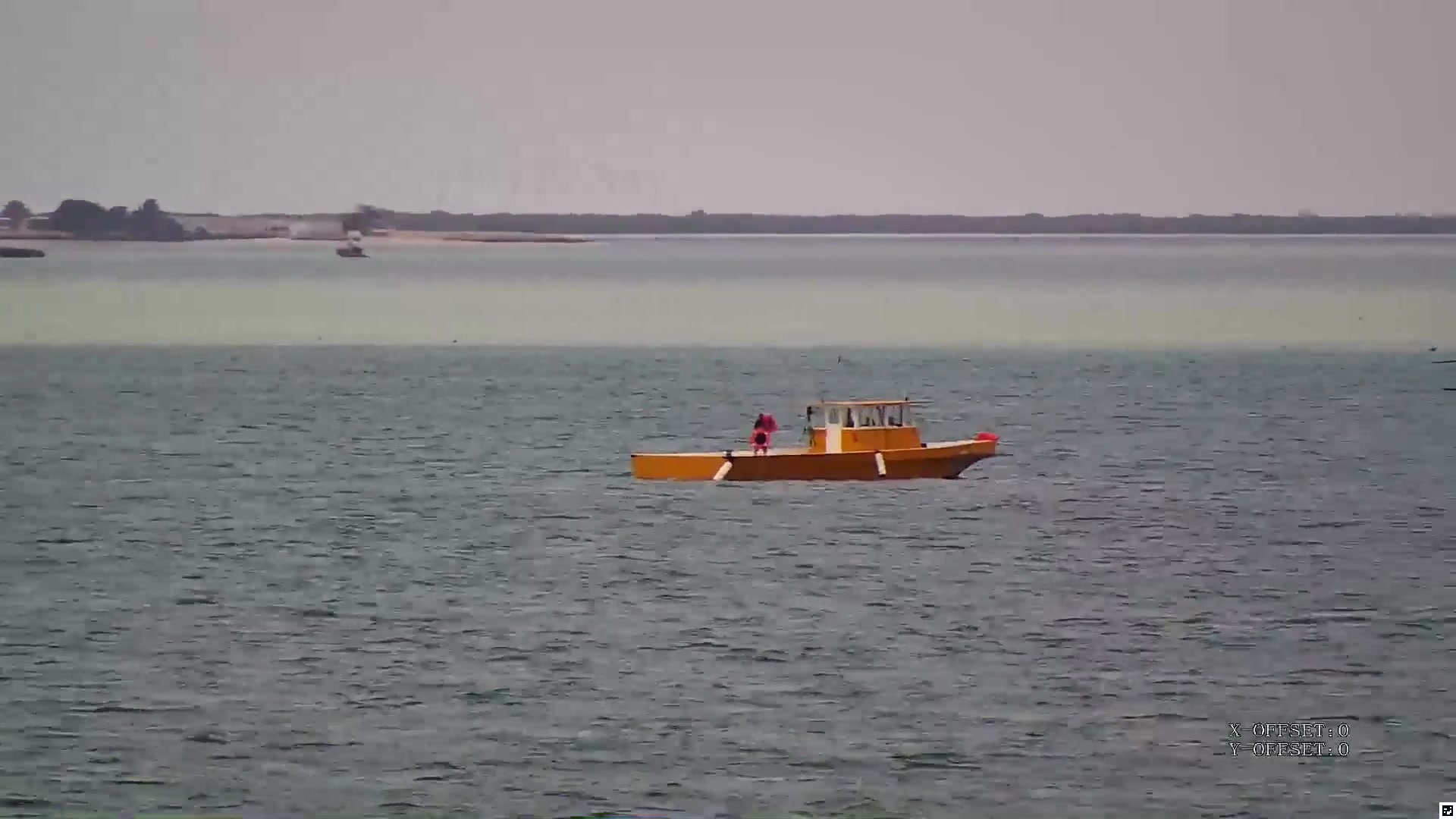}}
\hfill
\subfloat[USV acquisition.]{
    \includegraphics[width=0.235\textwidth]{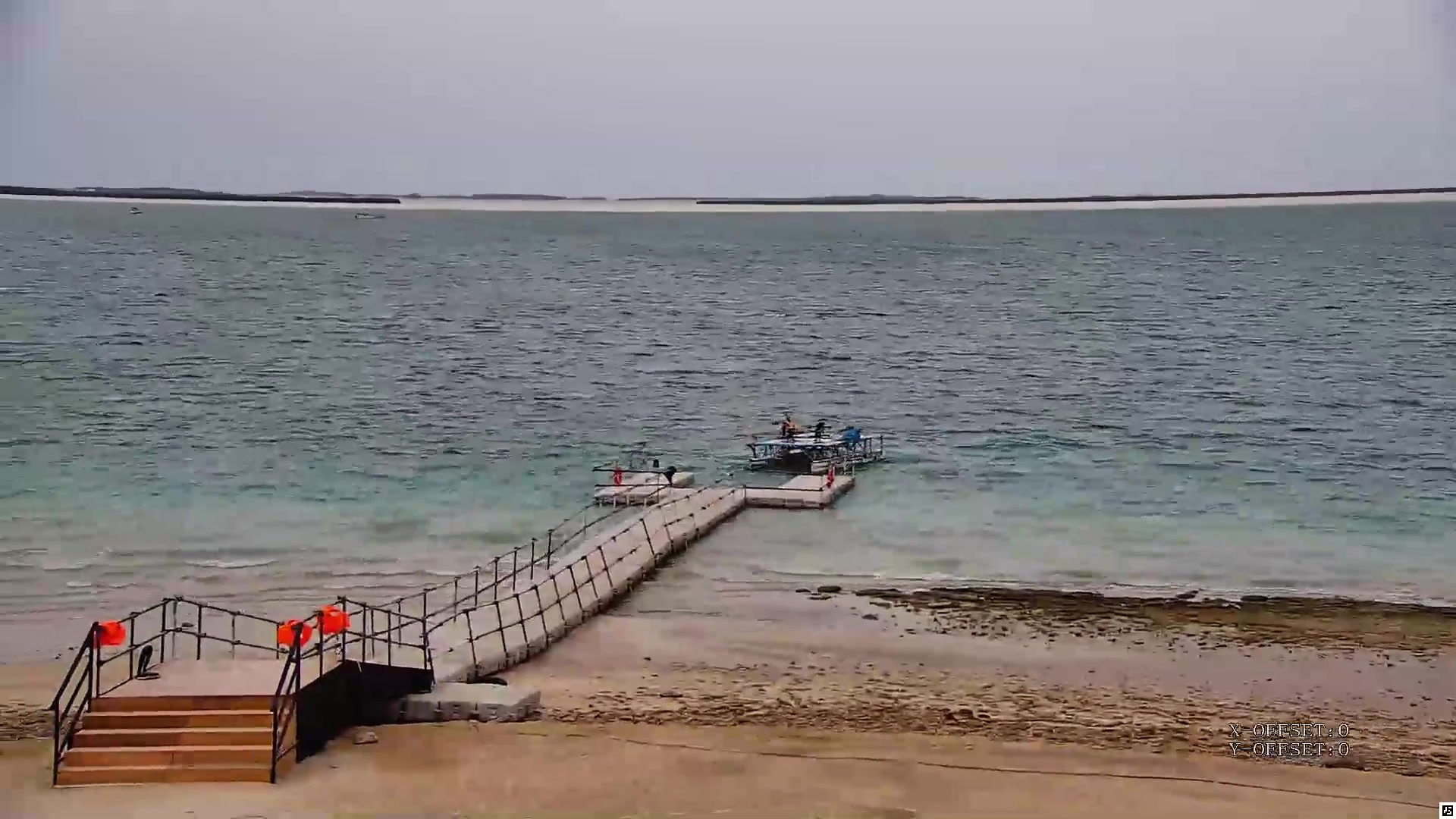}}
\\[-0.25em]
\subfloat[Navigation update.]{
    \includegraphics[width=0.235\textwidth]{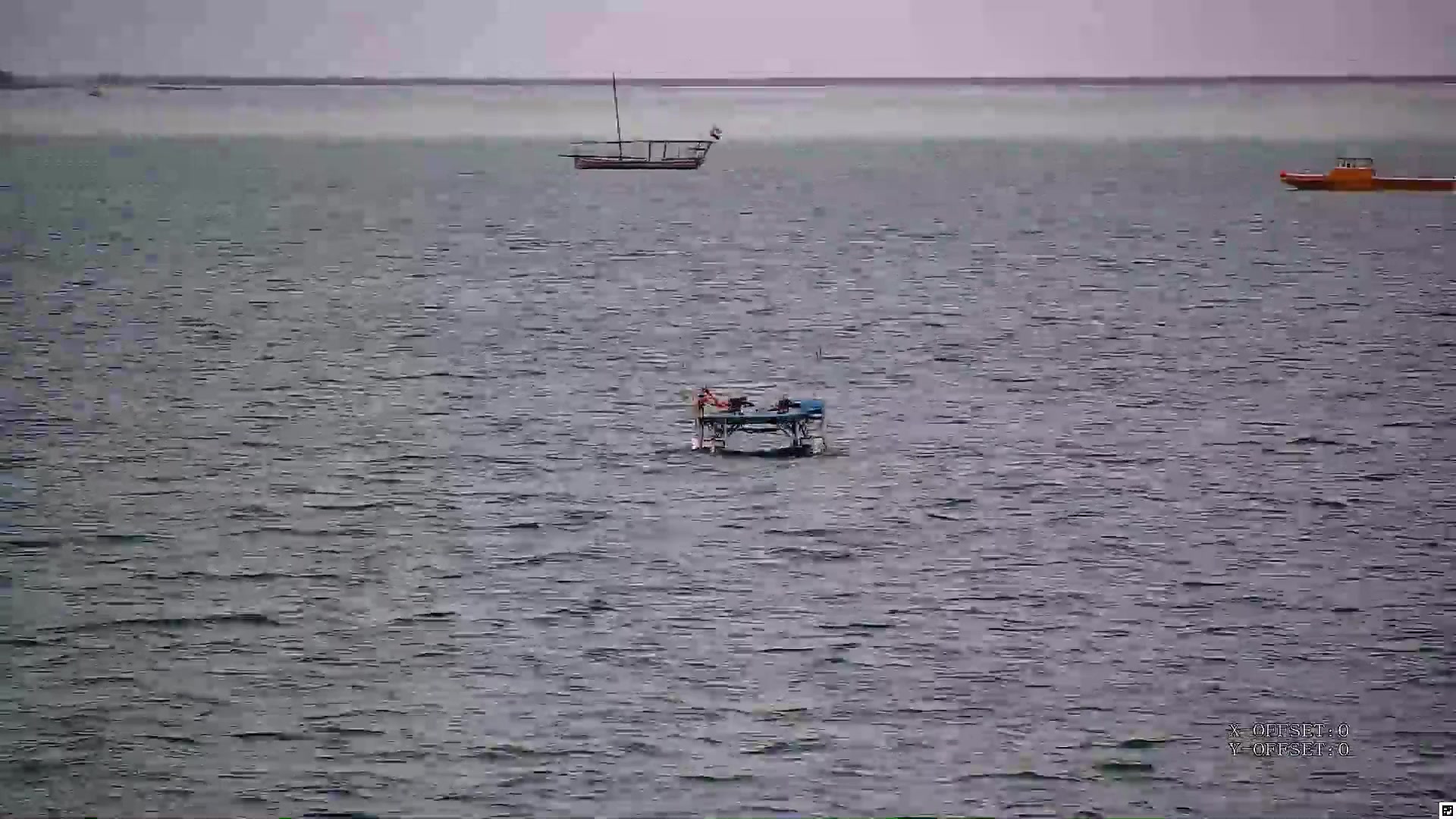}}
\hfill
\subfloat[Approach.]{
    \includegraphics[width=0.235\textwidth]{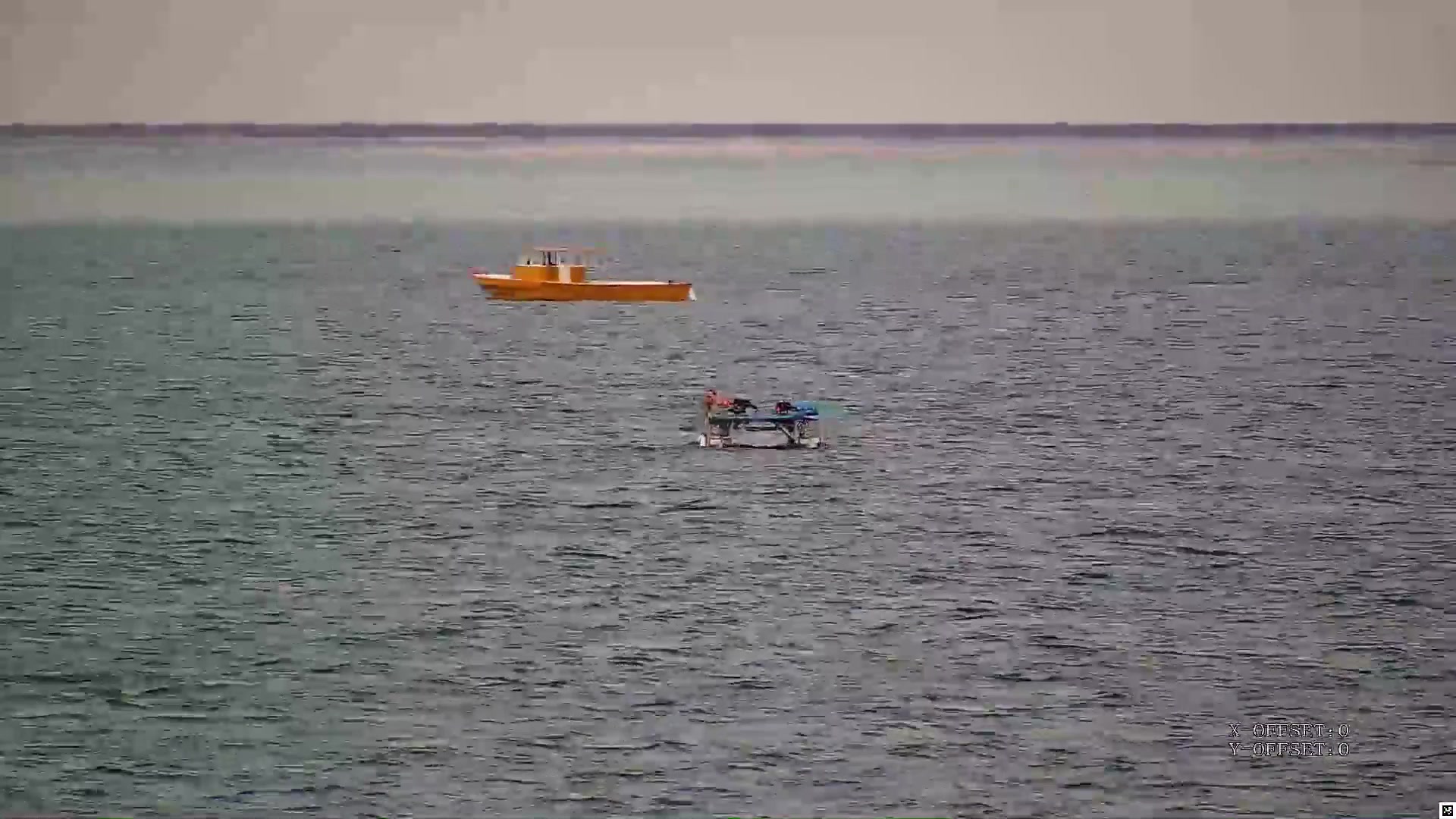}}
\hfill
\subfloat[Recovery scan.]{
    \includegraphics[width=0.235\textwidth]{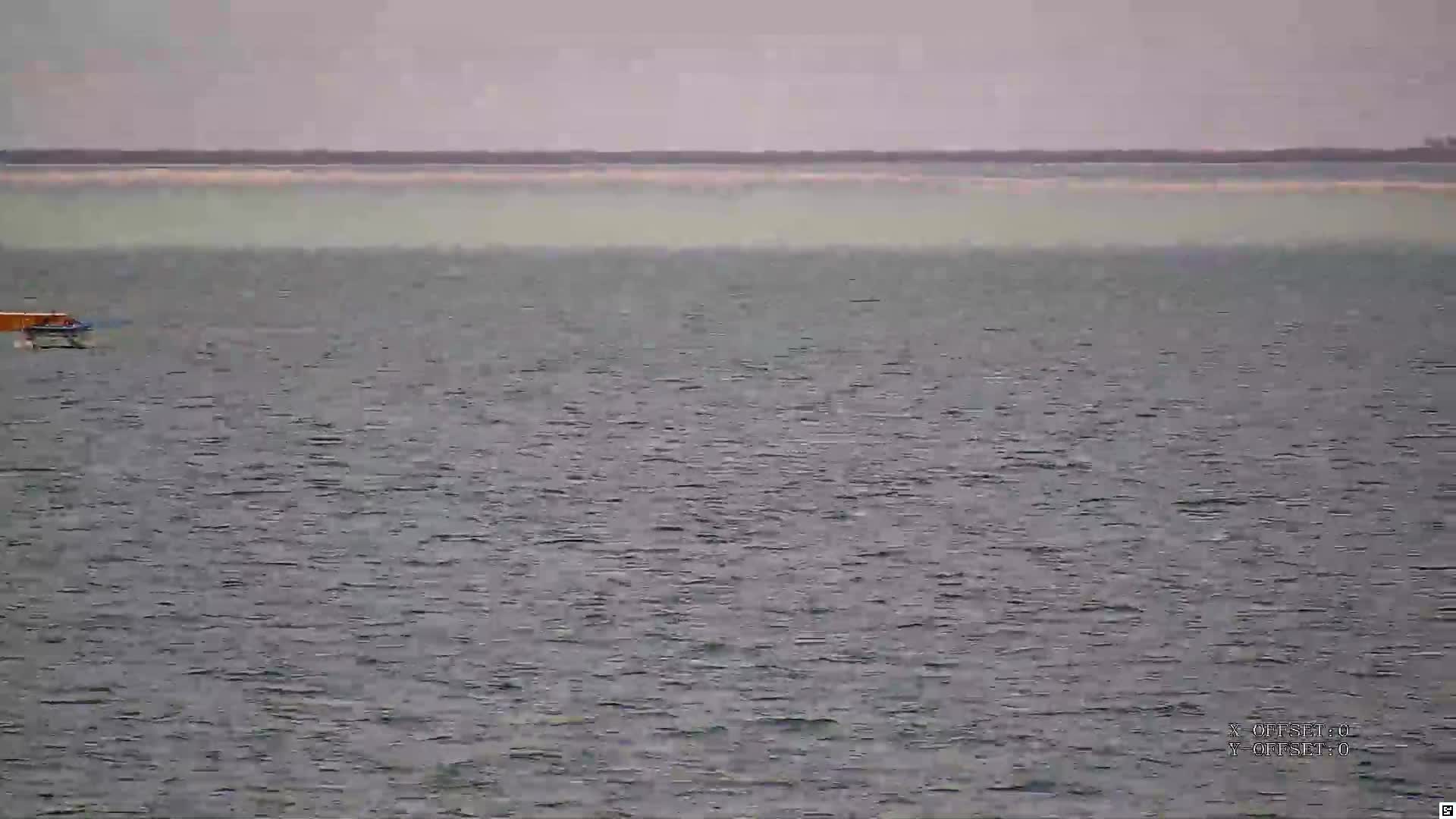}}
\hfill
\subfloat[Reacquisition.]{
    \includegraphics[width=0.235\textwidth]{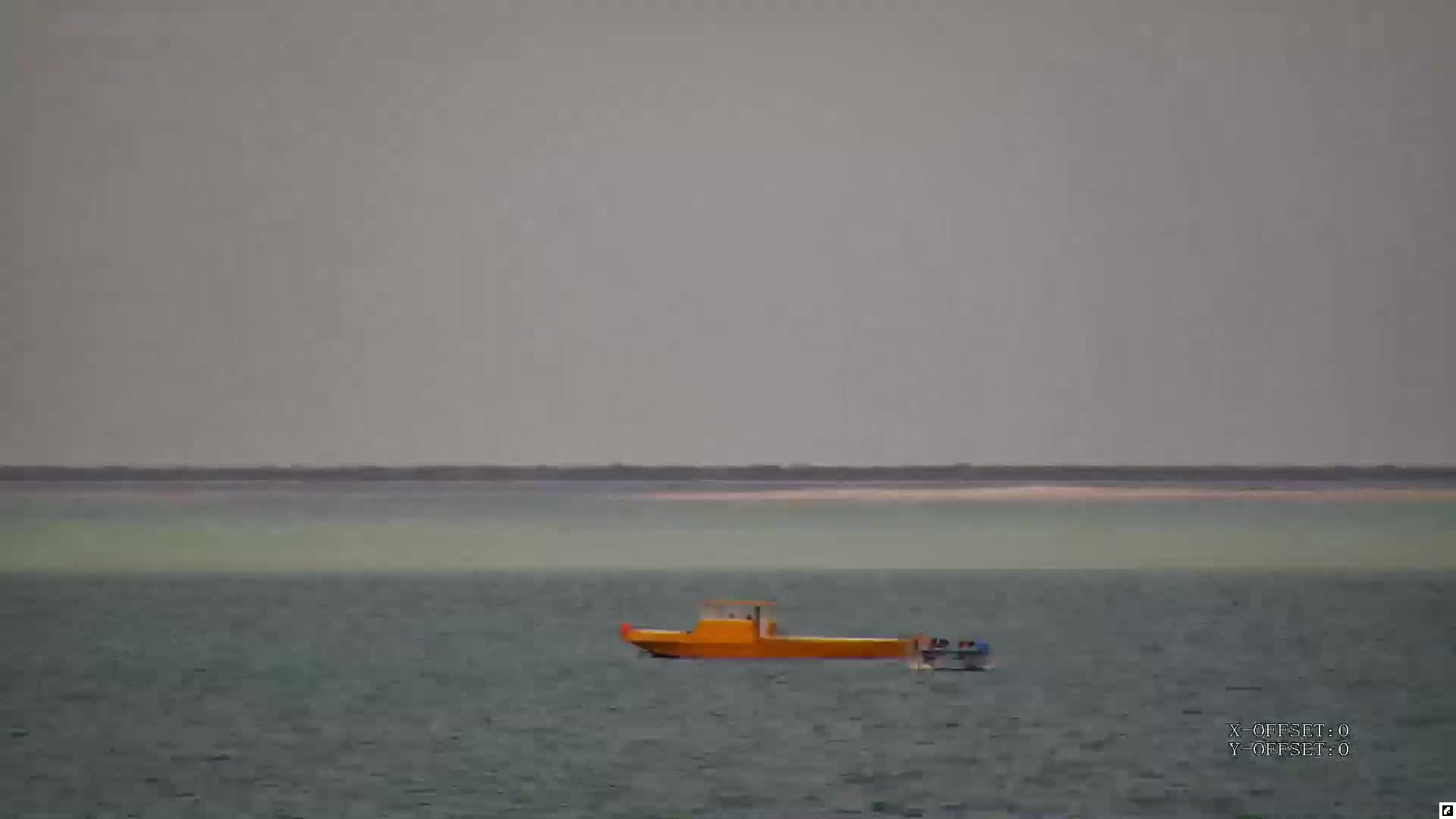}}
\caption{Representative onboard frames from the Yas Island mission.
The sequence shows the far-range search sweep,
target entry and confirmation,
cooperative-USV acquisition,
navigation updates,
the deliberately activated recovery interval,
and visual reacquisition.}
\label{fig:yas_video_sequence}
\end{figure*}

\begin{figure*}[!t]
\centering
\subfloat[Pod pointing trajectory.]{
    \includegraphics[width=0.31\textwidth]{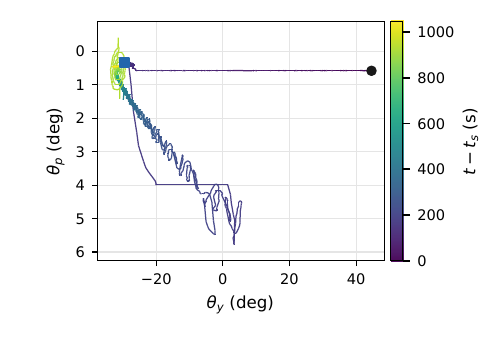}}
\hfill
\subfloat[Yaw and pitch histories.]{
    \includegraphics[width=0.31\textwidth]{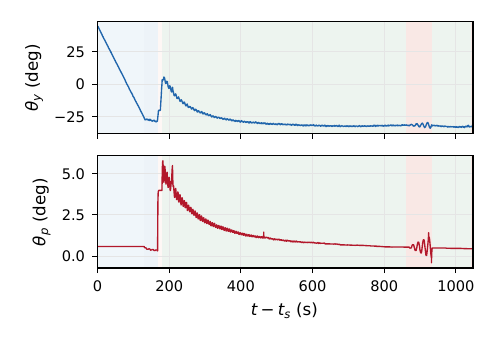}}
\hfill
\subfloat[HFOV and datalink range.]{
    \includegraphics[width=0.31\textwidth]{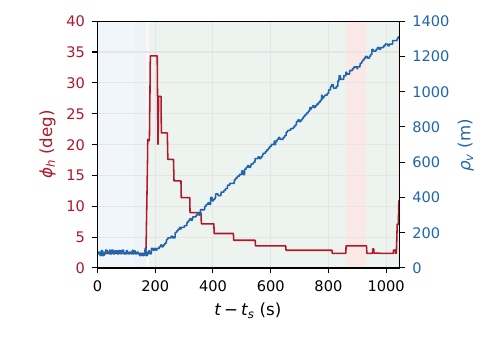}}
\caption{Synchronized pod and datalink histories during the integrated mission.
Background bands denote the supervisory states,
linking pod motion and FOV changes to search,
tracking,
guidance,
and recovery.}
\label{fig:yas_field_overview}
\end{figure*}

The integrated field mission was conducted during the final Yas Island
maritime deployment.
The logged data from this run support system-level evaluation of phase transitions
and target-relative navigation behavior.
Figs.~\ref{fig:yas_video_sequence} and \ref{fig:yas_field_overview}
summarize the onboard recognition sequence
and the synchronized pod and datalink histories.
The target was detected during the first far-range sweep;
because of this early detection,
the remaining planned near-range scan was not executed,
and the system automatically switched from search to tracking.

The mission transitions from search to tracking,
then to target confirmation,
and then to navigation.
During the available navigation window,
the datalink range increases as the USV moves away from the UAV,
whereas the estimated USV-to-target distance decreased from
\SI{1388.3}{\meter} to \SI{213.8}{\meter}.
The minimum UAV-side estimated target-relative distance over the available data is
\SI{167.6}{\meter}.
These online outputs characterize integrated navigation behavior
from UAV-side estimates.

\subsubsection{Visual-Loss Recovery}

\begin{figure}[!htbp]
\centering
\includegraphics[width=0.78\linewidth]{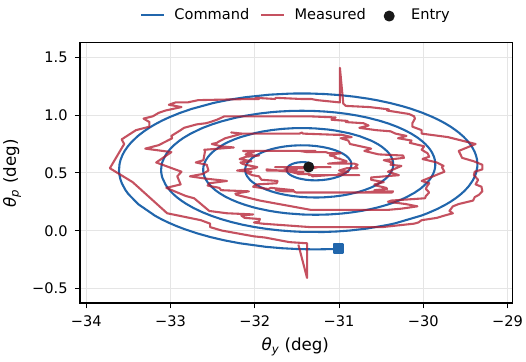}
\caption{Deliberately activated visual-loss recovery interval in pitch-yaw space.
The commanded expanding spiral and measured pod response are shown until reacquisition;
this single interval demonstrates the implemented transition but does not estimate a recovery probability.}
\label{fig:yas_recovery_detail}
\end{figure}

Fig.~\ref{fig:yas_recovery_detail} shows one deliberately triggered
visual-loss recovery interval in the mission data.
In this active recovery test,
the USV visual detection was intentionally allowed to remain unavailable
long enough to trigger local visual-loss recovery.
At the beginning of recovery,
the controller stores the last valid pod direction,
expands the horizontal FOV,
and performs a local pitch-yaw scan around that direction.
After reacquisition,
the system returns to navigation and resumes target-relative navigation updates,
showing the intended recovery-to-navigation transition.

\subsubsection{GPS-Referenced Localization Analysis}

\begin{figure}[!htbp]
\centering
\subfloat[Position error.]{
    \includegraphics[width=0.82\linewidth]{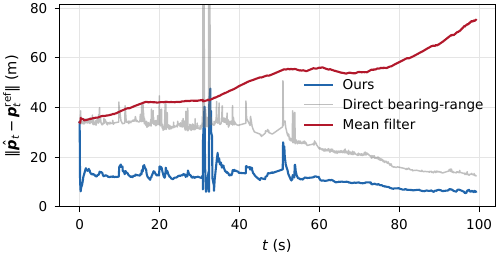}}
\vspace{0.2em}

\subfloat[Planar trajectory.]{
    \includegraphics[width=0.82\linewidth]{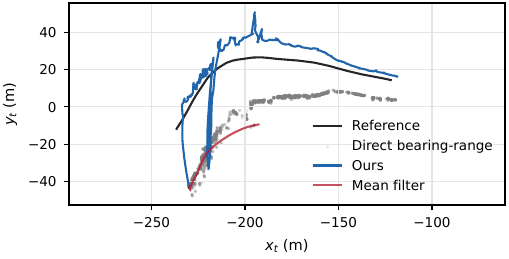}}
\caption{GPS-referenced field localization under a common synchronized measurement sequence.
(a) Position errors of AUSLUN,
direct bearing-range conversion,
and a mean filter;
the direct-conversion curve is clipped at its 99th percentile for display only.
(b) Corresponding planar trajectories.
AUSLUN achieves \SI{11.4}{\meter} mean error and \SI{16.0}{\meter} 95th-percentile error.}
\label{fig:auslun_estimator_field}
\end{figure}

The localization experiment uses a separate GPS-referenced Yas-region segment
to isolate estimator behavior from the integrated mission transitions.
AUSLUN is compared with two controlled non-recursive baselines
using exactly the same synchronized visual bearing,
datalink range,
UAV pose,
and reference trajectory.
The direct bearing-range baseline applies instantaneous geometric conversion,
whereas the mean-filter baseline averages these direct estimates.
The comparison therefore tests temporal consistency under matched measurements;
it is not a cross-platform comparison with radar,
LiDAR,
or onboard VIO systems,
whose sensing ranges and deployment conditions differ
\cite{liu23vio,han2019coastal,shen2023lidar,volden2022vision}.
The AUSLUN estimator is initialized by the first valid bearing-range measurement
and then applies the six-state recursive model in
\eqref{eq:state_vector}--\eqref{eq:estimator_update}.
Fig.~\ref{fig:auslun_estimator_field} summarizes the estimator error
and planar trajectory over the aligned field-data window.
Its mean error is \SI{11.4}{\meter}
and its 95th-percentile error is \SI{16.0}{\meter},
whereas the direct bearing-range and mean-filter baselines have mean errors of
\SI{27.5}{\meter} and \SI{50.1}{\meter},
respectively.
Thus,
the recursive estimator reduces mean error by 58.5\% relative to direct conversion
over this segment.

\section{Discussion}
\label{sec:discussion}

The experiments support the central systems claim:
a coastal fixed-hover UAV can extend the sensing horizon of a GNSS-denied USV
without requiring the UAV to execute an over-water coverage trajectory.
The benefit comes from separating platform stabilization from sensing motion.
VIO maintains the aerial reference near the shoreline,
whereas the pod supplies the angular motion and zoom required for search and tracking.
The planning results further show that polygon-dependent yaw bounds matter most
for narrow or irregular regions;
the improvement is smaller when the mission region already occupies most of the admissible sector.

The same fixed viewpoint also defines the main limitations.
Coverage completeness is established under the planar footprint and nominal target-size model,
not through an empirical probability-of-detection study.
Localization accuracy depends on VIO pose quality,
pod calibration,
visual centering,
range availability,
and the \SI{10}{\meter} datalink quantization used in the field system.
The present comparison isolates recursive estimation from two non-recursive alternatives,
but does not determine which individual error source dominates.
A calibration study,
range-quantization ablation,
and uncertainty propagation from localization to guidance
would strengthen the causal evidence for the estimator design.

Finally,
the integrated mission demonstrates supervisory feasibility rather than repeatability.
One deliberate interruption was sufficient to verify the implemented transition
from guidance to local recovery and back,
but repeated trials with controlled loss duration,
sea state,
target motion,
and background traffic are required to estimate recovery probability and mission completion rate.
The current architecture is therefore best interpreted as a field-validated baseline
for stationary-target coastal approach,
not as a general solution to moving-target pursuit or collision-aware maritime autonomy.

\section{Conclusion}
\label{sec:conclusion}

This paper presented AUSLUN,
a fixed-hover UAV--USV system that transfers search motion to a zoom pod
and closes the loop from maritime perception to target-relative guidance.
The polygon-aware planner reduced planned scan time by 10.9--55.6\%
relative to a matched fixed-sector scan,
and the modality-aware recursive estimator achieved
\SI{11.4}{\meter} mean error in the GPS-referenced field segment.
The integrated mission further demonstrated automatic transitions
from far-range search to target localization,
USV guidance,
local visual-loss recovery,
and navigation resumption.
These results support fixed-hover aerial assistance as a practical operating mode
when a coastal VIO-feasible viewpoint is available.

The demonstrated scope remains bounded.
The target was stationary,
the UAV maintained line of sight from one coastal viewpoint,
the integrated mission and recovery evidence were single-run demonstrations,
and collision avoidance,
final docking,
and close-range perception remained onboard USV functions.
Future work should evaluate repeated recovery trials,
moving targets,
uncertainty-aware guidance,
and active or multi-UAV viewpoint selection.

\section*{Acknowledgments}
This work was supported by Beijing Institute of Technology through the National Key Research and Development Program
under Grant No. 2022YFE0204400.

\bibliographystyle{IEEEtran}
\bibliography{ref}

\end{document}